\pdfoutput=1
\documentclass[10pt,twocolumn,letterpaper]{article}

\usepackage[pagenumbers]{cvpr}      %

\usepackage{cuted} %

\usepackage{currfile} %

\usepackage{caption} %

\usepackage{microtype}

\renewcommand{\paragraph}[1]{\vspace{.32em}\noindent\textbf{#1}\hspace{0.45em}\ignorespaces}

\usepackage{amsmath}
\usepackage{amssymb}
\usepackage{booktabs}
\usepackage{colortbl}
\usepackage{multirow}
\usepackage{xspace}
\definecolor{causalbg}{HTML}{DBEAFE}
\definecolor{decbg}{HTML}{F1F5F9}
\definecolor{harmbg}{HTML}{FEE2E2}
\definecolor{causaltext}{HTML}{1D4ED8}
\definecolor{dectext}{HTML}{334155}
\definecolor{harmtext}{HTML}{B91C1C}
\definecolor{oursbg}{HTML}{EFF6FF}
\definecolor{mutedgray}{HTML}{64748B}
\newcommand{\tm}{ThinkMorph\xspace}
\newcommand{\wan}{Wan-R1\xspace}
\newcommand{\vrlvr}{VideoRLVR\xspace}
\newcommand{\zebra}{Bagel-Zebra-CoT\xspace}
\newcommand{\canvas}{BAGEL-Canvas\xspace}
\newcommand{\vrb}{VR-Bench\xspace}
\newcommand{\vsp}{VSP\xspace}
\newcommand{\full}{\textsc{full}\xspace}
\newcommand{\fullp}{\textsc{full}$'$\xspace}
\newcommand{\blind}{\textsc{blind}\xspace}
\newcommand{\draft}[1]{\textsc{draft}$_{#1}$\xspace}
\newcommand{\stopm}[1]{stop@#1\xspace}

\definecolor{cvprblue}{rgb}{0.21,0.49,0.74}
\usepackage[pagebackref,breaklinks,colorlinks,allcolors=cvprblue]{hyperref}

\title{Sketch, Don't Paint: Allocating Test-Time Compute in Generative Visual Reasoning}

\author{Sheng Lu \qquad Guangpin Tao \qquad Tong Lu\thanks{Corresponding author.} \qquad Ying Tai\\
State Key Laboratory for Novel Software Technology, Nanjing University\\
{\tt\small \{shenglu,tgpin\}@smail.nju.edu.cn \qquad \{lutong,yingtai\}@nju.edu.cn}}

\begin{document}
\setcounter{footnote}{1}
\maketitle
\begin{strip}
    \centering
    \vspace{-2.5em}
    \includegraphics[width=\textwidth]{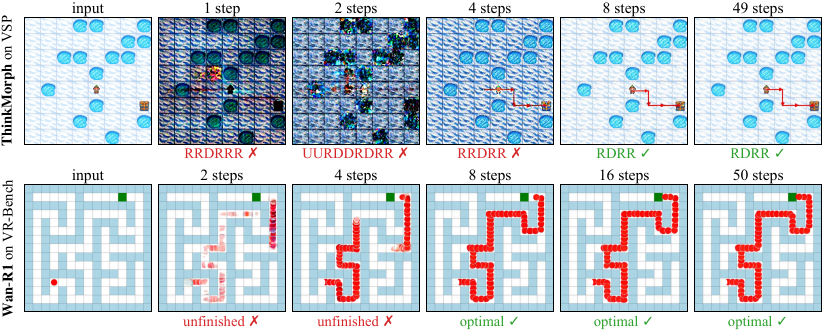}
    \captionof{figure}{%
        \textbf{Visual thoughts become legible long before they look photorealistic.}
        \textbf{Top:} intermediate thought images drawn by \tm on a \vsp puzzle from the same initial noise across $1$ to $49$ denoising steps, paired with the decoded action sequence (\textbf{U}p, \textbf{D}own, \textbf{L}eft, \textbf{R}ight).
        \textbf{Bottom:} 193-frame rollouts generated by \wan on a \vrb maze across $2$ to $50$ steps, shown as the overlaid trail of the ball.
        By $8$ steps, both models already lock in the plan of the full render, and later steps only refine texture. \emph{Sketch-and-vote} therefore votes over several 8-step sketches from independent noises and reaches higher accuracy at a lower budget (\cref{fig:budget}).
    }
    \label{fig:teaser}
    \vspace{0.4em}
\end{strip}

\begin{abstract}
Multimodal models increasingly reason by synthesizing pixels, sketching maze paths within interleaved chains of thought or rolling out video solutions frame by frame.
Each visual thought runs the full diffusion schedule inherited from photorealistic synthesis, so iterative denoising dominates inference cost.
We study how to allocate test-time compute along two axes: \emph{when} to render visual thoughts rather than thinking in text or answering directly, and \emph{how many} denoising steps each thought requires.
Controlled interventions that vary only how the thought is rendered uncover three regimes.
\emph{Causal} thoughts carry the plan and become legible early: an 8-step sketch nearly matches a 50-step render in accuracy, commits to the same plan as often as two independent full renders agree, and in video models mostly fails by stalling short of the goal, so the rollout abstains.
\emph{Decorative} thoughts can be omitted without loss, whereas \emph{harmful} thoughts lower accuracy.
Because the regime depends on both model and task, always drawing is suboptimal. Across 14 model-benchmark pairs of three interleaved reasoners, answering directly, without any thinking, wins 9 pairs, and choosing among direct, text-only, and visual tiers from ten labeled examples per pair raises mean accuracy from 60.7\% (always drawing) to 65.1\%.
Where drawing pays, we propose \emph{sketch-and-vote}, which samples 8-step sketches (a step count chosen without labels) from independent noises until one answer reaches three votes.
With this one setting, sketch-and-vote beats full rendering with fewer denoising steps on an image reasoner and two video reasoners, by 8.6 points on \vsp spatial planning and by 9.6 and 22.1 points on mazes.
\end{abstract}

\section{Introduction}
\label{sec:intro}

A growing class of multimodal models reasons by generating pixels.
Unified image-text models interleave textual chains of thought with synthesized images, sketching paths onto mazes, highlighting chart regions, or rearranging jigsaw pieces before answering~\cite{li2025mvot,chern2025thinking,li2025zebracot,gu2025thinkmorph,xu2025visualplanning}.
Video models enact the solution itself in frame space, \eg, rolling a ball through a maze~\cite{wiedemer2025video,tong2026thinking,yang2025vrbench}.
In both cases, every visual thought runs a dense denoising schedule inherited from photorealistic synthesis. For \tm~\cite{gu2025thinkmorph}, one 49-step thought image takes over half of spatial-planning latency, and for \wan~\cite{yang2025vrbench}, a 50-step, 193-frame rollout takes nearly three minutes on a single GPU.

A thought only has to be \emph{legible} for reasoning, and denoising steps spent on photographic detail beyond that point waste test-time compute.
Given a budget, should a model \emph{paint} one high-fidelity thought, \emph{sketch} several rough ones, think in text alone, or answer immediately?

\begin{figure*}[t]
    \centering
    \includegraphics[width=\textwidth]{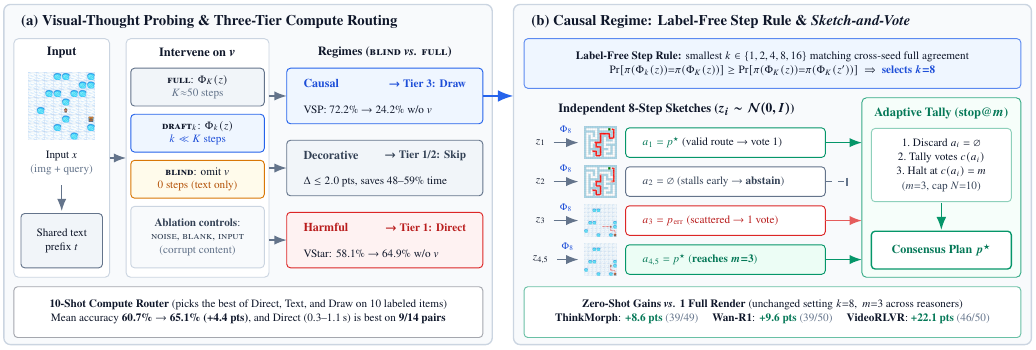}
    \caption{\textbf{Probing protocol and two-stage compute allocation.}
    \textbf{(a)} Holding $x$ and $t$ fixed, rendering interventions separate \emph{causal}, \emph{decorative}, and \emph{harmful} thoughts, motivating a 10-shot router across Direct, Text, and Draw tiers.
    \textbf{(b)} Where thoughts are causal, unlabeled commitment picks $k{=}8$ and \emph{sketch-and-vote} aggregates independent $k$-step sketches with self-abstention and early stopping (\stopm{$m$}).}
    \label{fig:overview}
\end{figure*}

We study this question with controlled interventions on the visual thought (\cref{fig:overview}a).
Holding the model, input, and pre-thought text fixed, we vary only how the thought is rendered: truncating denoising from the same or a new noise seed, omitting the thought altogether, or substituting raw noise, a blank canvas, or the input image (\cref{sec:probe}).
Across six benchmarks with \tm and the maze and Sokoban tasks of \vrb with \wan, visual thoughts fall into three regimes (\cref{tab:regimes}).
\emph{Causal} thoughts carry the solution: on \vsp spatial planning, omitting the thought drops \tm accuracy from 72.2\% to 24.2\%.
\emph{Decorative} thoughts accompany an answer already decided in text: on VisPuzzle, BLINK-Jigsaw, MMVP, and ChartQA, omitting them shifts accuracy by at most 2.0 points while halving latency.
\emph{Harmful} thoughts mislead: on VStar, omitting them \emph{raises} accuracy from 58.1\% to 64.9\%.

Which regime holds depends on both model and task, and the cheapest tier often scores highest. Direct answering takes at most 1.1\,s per question on \tm (\vs 17--73\,s for thinking) and wins five of its seven benchmarks, whereas \zebra~\cite{li2025zebracot} depends causally on its visual thoughts on VStar.
Choosing among direct, text-only, and visual tiers for each model-benchmark pair from ten labeled questions (\cref{sec:exp_tiers}) raises mean accuracy over 14 pairs of three interleaved reasoners from 60.7\% (always drawing) to 65.1\%.

Where visual thoughts are causal, coarse-to-fine flow integration locks in the plan early (\cref{fig:teaser,fig:commit}): an 8-step sketch nearly matches the same-seed full render in accuracy and commits to the same plan as often as two independent full renders agree (\eg, 65.8\% \vs 61.2\% on \vsp).
Sketch failures also favor voting: a failed video sketch usually stalls or dissolves before the goal and so abstains, and wrong paths in \tm's image sketches scatter across trajectories while correct ones agree.
We exploit both properties in \emph{sketch-and-vote} (\cref{fig:overview}b), which draws $k$-step sketches from independent noises, ignores abstentions, and stops once any answer reaches $m$ votes~\cite{wang2023selfconsistency,aggarwal2023adaptive}, where $k{=}8$ is chosen without labels as the fewest steps matching cross-seed full-render agreement.

With one fixed setting ($k{=}8$, $m{=}3$), sketch-and-vote beats full rendering at a lower step budget across image and video reasoners (\cref{tab:main,fig:budget}).
On \vsp, \tm reaches 80.3\% in 39 steps \vs 71.7\% in 49 steps (+8.6 points).
On \vrb mazes, \wan solves 84.0\% optimally in 39 steps \vs 74.4\% in 50 steps (+9.6 points). It comes close to a 10-draft ground-truth oracle (84.7\%), beats a 5-full-video vote (80.3\% at 250 steps), and does not trail full rendering on any of the 72 levels.
Transferred unchanged to the RL-trained reasoner \vrlvr~\cite{zhu2026video}, sketch-and-vote solves 86.8\% of mazes in 46 steps \vs 64.7\% for one full video (+22.1 points) and 79.0\% for a 3-full-video vote at 150 steps ($p{=}0.005$). On \vrlvr's SFT checkpoint, sketching at test time beats a full rollout of the RL model by 23.8 points (\cref{fig:analysis}c).
Gains are largest on the hardest instances: +22.3 points on $8{\times}8$ grids (69.3\% \vs 47.0\%, \cref{fig:analysis}a).
In wall-clock time, $m{=}2$ adds at most 6\% over one full render while keeping +7.0 and +9.6 points on \vsp and \wan mazes, and on \vrlvr stopping at the first valid sketch is faster than one full video (72\,s \vs 99\,s) and 20.1 points more accurate.
Voting saturates only when errors repeat across seeds, as when \wan repeats illegal pushes on Sokoban. There, sketch-and-vote matches full rendering in 32 \vs 50 steps, and a solution-free rule check restores a significant gain (\cref{fig:analysis}b).

In summary, our contributions are as follows:
\begin{itemize}[nosep]
    \item A controlled intervention protocol separating causal, decorative, and harmful visual thoughts, paired with a 10-shot three-tier router (+4.4 points over always drawing).
    \item Evidence that plans form early in denoising: an 8-step sketch is as plan-consistent as a full render, and incomplete video sketches self-abstain.
    \item \emph{Sketch-and-vote}, a training-free allocation rule with label-free step selection that beats full rendering at lower step budgets across image and video reasoners.
\end{itemize}

\section{Related Work}
\label{sec:related}

\paragraph{Reasoning with generated images.}
Visual intermediates aid multimodal reasoning whether produced via external code and vision tools~\cite{hu2024sketchpad,menon2024whiteboard} or structured edits~\cite{fu2025refocus}.
Recent models synthesize such images directly: MVoT~\cite{li2025mvot} renders spatial states step by step, Thinking with Generated Images~\cite{chern2025thinking} critiques and refines its own generations, Visual Planning~\cite{xu2025visualplanning} plans purely in pixel space, and Zebra-CoT~\cite{li2025zebracot}, MathCanvas~\cite{shi2025mathcanvas}, and \tm~\cite{gu2025thinkmorph} fine-tune unified backbones~\cite{deng2025bagel} on interleaved text-image traces.
Alternative designs reason in continuous latent tokens~\cite{yang2026mirage,li2026lvr,zhang2025latentsketchpad} or solve mazes and \vsp inside a single diffusion trajectory~\cite{dai2026endocot}.
Whereas prior work compares interleaved and text-only modes across distinct checkpoints or prompts, we hold the model and pre-thought text fixed and perturb only visual rendering, akin to truncation probes of textual chain-of-thought faithfulness~\cite{lanham2023faithfulness}.
While \tm notes that a mixed-mode variant scores higher in text than interleaved mode on MMVP and VStar~\cite{gu2025thinkmorph}, our interventions isolate the rendered image's causal effect and turn it into a compute routing rule.

\paragraph{Reasoning via video generation.}
Video generators solve spatial tasks directly in frame space, both zero-shot at frontier scale~\cite{wiedemer2025video,tong2026thinking} and after fine-tuning in open weights~\cite{yang2025vrbench,zhu2026video}.
\vrb~\cite{yang2025vrbench} adapts Wan2.2~\cite{wan2025wan} into \wan via task LoRAs (later trained with RL under the same name~\cite{liu2026wanr1}), while \vrlvr~\cite{zhu2026video} optimizes Wan2.2 with verifiable rewards. We evaluate the released supervised \wan checkpoints alongside \vrlvr's RL and SFT models.
Prior studies note that sampling extra seeds raises oracle Pass@$K$~\cite{yang2025vrbench} and that voting over five full Sora-2 rollouts improves puzzle accuracy~\cite{tong2026thinking}.
Both run every rollout at full fidelity, whereas we select solutions without ground truth and show that spending these steps on cheap sketches instead works better.

\paragraph{Test-time compute.}
Self-consistency~\cite{wang2023selfconsistency} votes across sampled chains of thought, and adaptive variants stop once consensus forms~\cite{aggarwal2023adaptive,li2024esc}. Compute-optimal scaling balances sample breadth against per-sample depth~\cite{snell2025scaling}, and text reasoners can often skip thinking with little loss~\cite{ma2025nothinking}.
In unified multimodal models, UniT~\cite{chen2026unit} favors sequential refinement over parallel full-fidelity sampling.
In diffusion models, extra compute typically drives verifier-guided search over initial noises~\cite{ma2025inference}, partial videos~\cite{liu2025videot1}, or partially denoised trajectories~\cite{yoon2025mctd} in continuous planners~\cite{janner2022diffuser}.
We instead trade denoising steps per sample for sample count to maximize the accuracy of a discrete answer, require no external verifier, and jointly decide when to skip drawing or thinking altogether.

\paragraph{Fast sampling and coarse-to-fine denoising.}
Fast ODE solvers~\cite{song2021ddim,lu2022dpmsolver}, straightened flows~\cite{lipman2023flow,liu2023rectified}, and distillation~\cite{song2023consistency} shorten sampling schedules for perceptual synthesis. Diffusion is known to set global structure early and fine detail late~\cite{choi2022p2,balaji2022ediffi,hertz2023prompt}, and dMLLM-TTS~\cite{xin2026dmllmtts} uses this to prune candidates by prompt alignment.
Our sketches only re-discretize each model's native sampler, with no retraining. Because the symbolic plan forms early, a few-step draft is already enough for reasoning.

\section{Probing and Allocating Visual Thoughts}
\label{sec:method}

\subsection{Generative visual reasoners}
\label{sec:setup}

We consider a multimodal model that receives an input $x$ (images and a question), writes an optional textual prefix $t$, and synthesizes a visual thought $v$ by integrating a learned flow for $K$ steps from Gaussian noise $z$:
\begin{equation}
    v = \Phi_K(z \mid x, t), \qquad z \sim \mathcal{N}(0, I).
\end{equation}
Each step runs one denoiser pass per classifier-free guidance branch over all latent image or video tokens.
In an interleaved reasoner such as \tm~\cite{gu2025thinkmorph}, the model re-encodes $v$ into its context and continues decoding to output an answer $a = R(x, t, v)$.
In a video reasoner such as \wan~\cite{yang2025vrbench} or \vrlvr~\cite{zhu2026video}, the rollout itself enacts the solution and a deterministic readout extracts $a = \pi(v)$ from the frames.
In both families, $K$ defaults to a photorealistic schedule: 49 Euler steps for \tm, and 50 steps for \wan and \vrlvr.

\subsection{Interventions on the thought}
\label{sec:probe}

To isolate the causal contribution of $v$, we hold $x$ and $t$ fixed and intervene only on how $v$ is rendered, leaving the sampler, guidance scales, visual decoder, and re-encoding pipeline unchanged (\cref{fig:overview}a):
\begin{itemize}[nosep]
    \item \full: the default $K$-step trajectory $\Phi_K(z \mid x, t)$.
    \item \draft{k}: the native sampler re-discretized into $k \ll K$ steps from the \emph{same} initial noise $z$, decoded and read identically to \full.
    \item \fullp: a $K$-step render from an independent noise draw $z'$, measuring intrinsic cross-seed variation.
    \item \blind: thought omission, where autoregressive text decoding resumes immediately without visual thought tokens.
    \item \textsc{noise}, \textsc{blank}, \textsc{input}: content-ablated controls replacing $v$ with raw noise $z$, a uniform gray canvas, or a copy of the input image.
\end{itemize}
Because video reasoners emit no post-thought text, only \full, \draft{k}, and \fullp apply to them.
Comparing \blind against \full via McNemar's test partitions thoughts into \emph{causal} (omission lowers accuracy), \emph{decorative} (no significant change), and \emph{harmful} (omission raises accuracy) regimes.

\subsection{Reading the plan off a thought}
\label{sec:readout}

\paragraph{Plan extraction and optimality.}
In spatial planning, the answer is a discrete trajectory: an action sequence on \vsp~\cite{wu2025vsp} and a cell route on \vrb and \vrlvr.
For \tm, we parse the action sequence from its final boxed answer.
For \wan, following \vrb, we track the agent with CSRT~\cite{lukezic2017csrt}, map positions to grid cells, prune jitter loops, and append the goal cell if the 193-frame rollout ends while the agent steps into the goal from an adjacent cell.
\vrb's official metrics mix route quality with other factors: Success Rate (SR) checks only terminal goal proximity and rewards wall-crossing trajectories (released Sora-2 videos score 75.0\% SR although only 23.6\% follow valid routes), whereas Exact Match (EM) enforces a 1\%-frame-width pixel tolerance against one reference rollout and rejects all but 3.7\% of our optimal full renders.
We therefore evaluate \emph{plan optimality}: a maze route must start at the agent, reach the goal without entering walls or traps, and match the shortest-path length.
On Sokoban, we track the box by color and score a box route as plan-optimal if it reaches the target in the fewest reachable, legal pushes.
On \vrlvr, which paints routes onto tree-shaped mazes rather than moving an agent, we detect blocks whose color shifts between the first and last frames. If at least half of painted blocks lie on the shortest painted start-to-goal path, that path defines the route (otherwise the video abstains) and is plan-optimal whenever wall-free.
Our metric scores 100\% of ground-truth reference videos as plan-optimal across tasks, whereas official SR rejects 25\% of ground-truth Sokoban videos and \vrlvr's check fails 15\% of its own ground truth.

\paragraph{Plan commitment and agreement.}
Two thoughts commit to the same plan if they yield identical action sequences (\vsp) or cell routes (\vrb agent or box paths regardless of goal arrival, and \vrlvr painted routes), while rollouts with no extractable plan commit to none.
We define the \emph{commitment} of \draft{k} as the fraction of questions where $\Phi_k(z)$ matches the plan of $\Phi_K(z)$ from the shared seed $z$, and compare it against cross-seed \emph{agreement} between independent full renders averaged over all full-render pairs.
Once commitment reaches cross-seed agreement, truncating to $k$ steps perturbs the plan no more than resampling initial noise.

\begin{table*}[t]
    \centering
    \footnotesize
    \begin{minipage}[c]{0.575\textwidth}
        \centering
        \textbf{(a) Full nine-intervention suite on core benchmarks} \\[0.25em]
        \setlength{\tabcolsep}{5.2pt}
        \begin{tabular}{@{}lcccc@{}}
            \toprule
            Thought variant & \vsp & VisPuzzle & BLINK-Jigsaw & MMVP \\
            \midrule
            \full (49 steps)           & 72.2 & 71.8 & 63.3 & 81.3 \\
            \fullp (new noise)         & 71.2 & 72.3 & 64.7 & 81.0 \\
            \draft{8} (same noise)     & 69.7 & 72.3 & 68.0 & 81.3 \\
            \draft{4} (same noise)     & 57.8$^\downarrow$ & 71.5 & 66.0 & 81.3 \\
            \draft{1} (same noise)     & 39.0$^\downarrow$ & 70.8 & 66.7 & 80.7 \\
            \midrule
            \blind (omit thought)      & 24.2$^\downarrow$ & 71.5 & 65.3 & 81.0 \\
            \textsc{blank} (gray)      & 23.5$^\downarrow$ & 70.0 & 66.7 & 79.3 \\
            \textsc{input} (copy)      & 27.3$^\downarrow$ & 71.0 & 64.7 & 82.3 \\
            \textsc{noise} (raw $z$)   & 20.3$^\downarrow$ & 69.0 & 59.3 & 73.0$^\downarrow$ \\
            \midrule
            Agree w/ \full: \fullp     & 61.2 & 96.5 & 93.3 & 98.3 \\
            Agree w/ \full: \blind     & 16.8 & 96.0 & 95.3 & 97.0 \\
            \bottomrule
        \end{tabular}
    \end{minipage}\hfill
    \begin{minipage}[c]{0.405\textwidth}
        \centering
        \includegraphics[width=\linewidth]{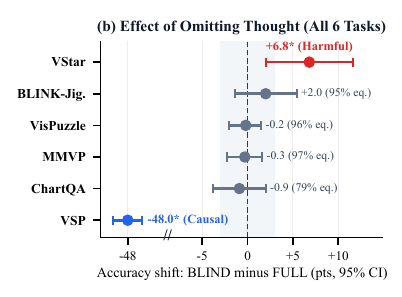}
    \end{minipage}
    \caption{\textbf{Controlled interventions on \tm's visual thoughts} (accuracy and agreement with \full, \%). \textbf{(a)} Complete nine-intervention matrix across four benchmarks ($p<0.05$ shifts marked by $\downarrow$, exact McNemar with Holm correction). \textbf{(b)} Accuracy shift of omitting the thought (\blind minus \full, 95\% CI and answer equivalence rate) across all six benchmarks, including VStar (\full 58.1\%, \draft{8} 64.4\%$^\uparrow$, \blind 64.9\%$^\uparrow$) and ChartQA (\full 75.7\%, \draft{8} 77.0\%, \blind 74.8\%).}
    \label{tab:regimes}
\end{table*}

\subsection{Three tiers of test-time compute}
\label{sec:tiers}

An interleaved reasoner can answer \emph{directly} without chain-of-thought prompting, think in \emph{text} alone (\blind), or \emph{draw} visual thoughts (\full, or sketch-and-vote).
We calibrate the tier per model and benchmark on $n$ labeled questions, picking the most accurate tier and breaking ties toward the cheaper option.

\subsection{Sketch-and-vote}
\label{sec:vote}

Where thoughts are causal, we draw $k$-step sketches $v_1, v_2, \dots$ from independent noises and read an answer $a_i$ off each (\cref{fig:overview}b).
A sketch self-abstains ($a_i = \varnothing$) when no complete plan emerges, namely when \wan's agent (or box) never reaches the goal or \vrlvr paints no valid path. This test needs no ground truth.
We take a majority vote over non-abstaining answers with uniform tie-breaking, failing only if all sketches abstain.
With early stopping (\stopm{$m$})~\cite{aggarwal2023adaptive}, sketches are drawn until one answer holds $m$ votes or $N{=}10$ sketches are drawn, costing $k$ times the number drawn.
In \tm, the prefix $t$ is encoded once and post-thought readings are decoded as one batch.

\paragraph{Hyperparameters and statistics.}
We select $k$ without labels as the smallest $k \in \{1, 2, 4, 8, 16\}$ whose plan commitment is not significantly below cross-seed full-render agreement (95\% paired bootstrap CI over questions).
The rule needs only unlabeled renders on a small sample of questions and picks $k{=}8$ for both \tm and \wan (\cref{fig:commit}, tested across all tasks).
We fix $(k{=}8, m{=}3, N{=}10)$ on \vsp and \vrb mazes (Style~1) and reuse it unchanged for all other styles, puzzles, and models.
Expected voting accuracy is computed over random subsets and permutations of each question's rendered pool and tested against mean full-render accuracy (two renders on \vsp, three to five on video tasks) via a two-sided paired permutation test ($10^4$ sign flips).

\section{Experiments}
\label{sec:experiments}

\subsection{Setup}
\label{sec:exp_setup}

\paragraph{Models.}
\tm~\cite{gu2025thinkmorph} fine-tunes BAGEL-7B-MoT~\cite{deng2025bagel} on interleaved text-image traces, generating thought images at input resolution with 49 Euler steps (flow shift 3, text/image guidance scales 4 and 2) and text temperature 0.3.
For tier routing, we also evaluate \zebra~\cite{li2025zebracot} and \canvas~\cite{shi2025mathcanvas}, which fine-tune the same backbone on general interleaved traces and geometric diagram reasoning under their authors' settings.
For video reasoning, \wan~\cite{yang2025vrbench} adapts Wan2.2-TI2V-5B~\cite{wan2025wan} with one LoRA per \vrb task (193 frames at $512{\times}512$, 50 steps, flow shift 5, guidance scale 5), while \vrlvr~\cite{zhu2026video} trains the same base model via RL from verifiable rewards (81 frames at $832{\times}480$, 50 steps).
All $k$-step sketches re-discretize each model's native schedule.

\paragraph{Benchmarks and compute.}
We probe \tm across six benchmarks: \vsp~\cite{wu2025vsp} navigation (600 mazes, 100 per size from $3{\times}3$ to $8{\times}8$, posed in \tm's native navigation format so it solves the test map rather than echoing VLMEvalKit's worked example), VisPuzzle (400 in-domain jigsaws), BLINK-Jigsaw (150)~\cite{fu2024blink}, MMVP (300)~\cite{tong2024mmvp}, VStar (191)~\cite{wu2024vstar}, and the 444 horizontal-bar ChartQA items~\cite{masry2022chartqa} curated by ReFocus~\cite{fu2025refocus}.
Tier routing adds HR-Bench 4K single-instance perception (400)~\cite{wang2025hrbench} for \tm, evaluates \zebra on the six benchmarks (240 stratified \vsp items), and tests \canvas on MathVista testmini geometry (208)~\cite{lu2024mathvista}.
For \wan, we run \vrb regular mazes and Sokoban (72 levels each: 24 easy, 24 medium, 24 hard) in Style~1, regular mazes in Style~2, and trap fields, which \wan saturates (100\% optimal at 50 steps and 98.6\% at 2 steps).
For \vrlvr, we evaluate 200 test mazes stratified by size (32 easy, 97 medium, 71 hard).
Interleaved models are scored deterministically with VLMEvalKit's rule-based matchers~\cite{duan2024vlmevalkit} as extended by \tm (valid goal-reaching paths on \vsp, option matching on multiple choice, and 5\% relative tolerance on ChartQA, which gives slightly lower absolute numbers than the GPT-judged scores in~\cite{gu2025thinkmorph}).
Videos are scored by plan optimality (\cref{sec:readout}) alongside official checks.
All runs use RTX~4090 GPUs. For each item we render \full, \fullp, the probes of \cref{sec:probe}, and ten independent 8-step sketches (five full videos per maze in \vrb Style~1 and three on Sokoban, Style~2, and \vrlvr).

\subsection{Three regimes of visual thoughts}
\label{sec:exp_regimes}

\Cref{tab:regimes} sorts \tm's tasks into the three regimes.
On \vsp, visual thoughts are \emph{causal}: ablating spatial content via omission, a blank canvas, repeating the input, or injecting raw noise drops accuracy by 45 to 52 points (\cref{tab:regimes}a).
On VisPuzzle, BLINK-Jigsaw, and MMVP, visual thoughts are \emph{decorative}.
Omitting them gives the same prediction as \full on 95--97\% of questions (matching cross-seed \full-\fullp agreement) and shifts accuracy by at most 2.0 points (95\% paired CIs exclude losses above 2.3 points) while saving 48--59\% of latency.
\tm states its answer in text \emph{before} drawing on 41\% of BLINK-Jigsaw and 28\% of MMVP items and keeps it 98\% of the time, even though it still attends to the image slot (raw noise lowers MMVP by 8.3 points).
ChartQA is likewise decorative (\cref{tab:regimes}b): \blind reaches 74.8\% \vs 75.7\% for \full ($p{=}0.65$), with six text-only seeds excluding losses beyond 2.9 points, and lower raw agreement on open-ended numbers (78.6\%) reflects text sampling variance (86\% across text-only runs).
Finally, on VStar, visual thoughts are \emph{harmful}: omitting them raises accuracy from 58.1\% to 64.9\% (95\% CI $[2.0, 11.6]$, and +5.5 points across six text-only seeds, $p{=}0.007$), while truncating to \draft{8} recovers nearly all of this gain (64.4\%).

\subsection{Choosing a tier from a few labels}
\label{sec:exp_tiers}

\begin{table}[t]
    \centering
    \footnotesize
    \setlength{\tabcolsep}{3.6pt}
    \begin{tabular}{@{}lrrrrr@{}}
        \toprule
        & \multicolumn{3}{c}{Always one tier} & \multicolumn{2}{c}{Tier per benchmark} \\
        \cmidrule(lr){2-4} \cmidrule(l){5-6}
        Model (benchmarks) & Direct & Text & Draw & Picked & Best \\
        \midrule
        \tm (7)          & 65.2 & 63.5 & 69.9 & \textbf{74.6} & 76.0 \\
        \zebra (6)       & \textbf{52.1} & 41.6 & 46.4 & 51.5 & 54.6 \\
        \canvas (1)      & 78.8 & \textbf{83.7} & 82.7 & 81.0 & 83.7 \\
        \midrule
        All 14 pairs     & 60.6 & 55.6 & 60.7 & \textbf{65.1} & 67.4 \\
        \bottomrule
    \end{tabular}
    \caption{\textbf{Routing across three tiers of test-time compute.} Mean accuracy (\%) across benchmarks when fixing one tier globally (Direct, Text, or Draw), picking a tier per benchmark from 10 labeled questions (scored on held-out items across 2000 random draws), or taking the per-benchmark oracle (\emph{Best}). Bold marks the strongest non-oracle strategy.}
    \label{tab:tiers}
\end{table}

\begin{figure}[t]
    \centering
    \includegraphics[width=\linewidth]{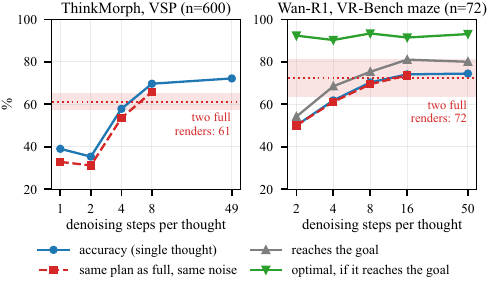}
    \caption{\textbf{Eight denoising steps make a sketch as plan-consistent as a second full render.} Single-thought accuracy and same-noise plan commitment across step counts $k$, plotted alongside cross-seed agreement between independent full renders (dotted line and 95\% band). Our label-free rule selects the smallest $k$ not significantly below cross-seed agreement ($k{=}8$).}
    \label{fig:commit}
\end{figure}

\begin{table*}[t]
    \centering
    \footnotesize
    \setlength{\tabcolsep}{5pt}
    \begin{tabular}{@{}l rr rrr rr rrr@{}}
        \toprule
        & \multicolumn{2}{c}{\tm, \vsp} & \multicolumn{3}{c}{\wan, maze} & \multicolumn{2}{c}{\wan, Sokoban} & \multicolumn{3}{c}{\vrlvr, maze} \\
        \cmidrule(lr){2-3} \cmidrule(lr){4-6} \cmidrule(lr){7-8} \cmidrule(l){9-11}
        Allocation of denoising steps & Acc. & Steps & Acc. & Steps & Seconds & Acc. & Steps & Acc. & Steps & Seconds \\
        \midrule
        One full render                       & 71.7 & 49   & 74.4 & 50  & 167 & 79.2 & 50 & 64.7 & 50 & 99 \\
        Two full renders, vote                & 71.8 & 98   & 77.7 & 100 & 333 & 80.0 & 100 & 73.4$^\dagger$ & 100 & 199 \\
        All full renders (3--5), vote         & \multicolumn{2}{c}{\textcolor{mutedgray}{\textit{pool = 2}}} & 80.3$^\dagger$ & 250 & 833 & \textbf{80.5}$^\ast$ & 150 & 79.0$^{\ast\dagger}$ & 150 & 298 \\
        Full renders, \stopm{1}               & 71.8 & 49   & 79.3$^\dagger$ & 76  & 252 & 80.2 & 52 & 77.3$^\dagger$ & 70 & 139 \\
        \midrule
        One 8-step draft                      & 68.4 & 8    & 70.3 & 8   & 48  & 74.3 & 8 & 58.6 & 8 & 29 \\
        8-step drafts, \stopm{1}              & 68.5 & 8.0  & 81.0$^\dagger$ & 16  & 95  & 76.5 & 9 & 84.8$^\dagger$ & 20 & 72 \\
        8-step drafts, \stopm{2}              & 78.7$^\dagger$ & 25.8 & \textbf{84.0}$^\dagger$ & 29 & 176 & 78.7 & 22 & 86.6$^\dagger$ & 35 & 127 \\
        \rowcolor{oursbg}
        8-step drafts, \stopm{3} (default)    & 80.3$^\dagger$ & 39.1 & \textbf{84.0}$^\dagger$ & 39 & 234 & 77.9 & 32 & \textbf{86.8}$^\dagger$ & 46 & 167 \\
        Ten 8-step drafts, vote               & \textbf{80.8}$^\dagger$ & 80 & \textbf{84.0}$^\dagger$ & 80 & 477 & 77.9 & 80 & \textbf{86.8}$^\dagger$ & 80 & 294 \\
        \bottomrule
    \end{tabular}
    \caption{\textbf{Sketch-and-vote \vs full rendering.} Accuracy (\%) and denoising steps per question across 1000 random draw orders (single renders average over all renders per question). \stopm{$m$} halts once one answer holds $m$ votes (up to ten sketches or all full renders), and the blue row marks our default setting ($k{=}8$, \stopm{3}). Seconds report single-GPU latency per maze. $^\dagger$: beats one full render at $p<0.01$ (paired permutation test). $^\ast$: pool of three full renders (\wan maze uses five, and \vsp uses two).}
    \label{tab:main}
\end{table*}

\begin{figure*}[t]
    \centering
    \includegraphics[width=\linewidth]{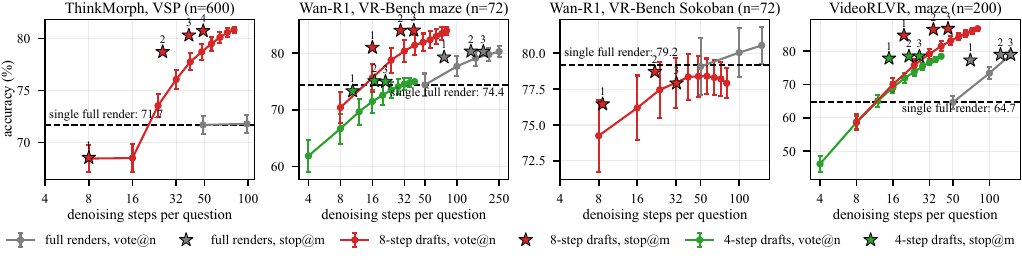}
    \caption{\textbf{Accuracy \vs denoising budget} (thoughts $\times$ steps per thought) for majority voting over $n$ thoughts (vote@$n$, mean $\pm$ s.d.\ over 1000 draws) and adaptive early stopping (\stopm{$m$}, stars labeled with $m$). On \vsp and both maze tasks, 8-step sketches surpass full rendering at a fraction of its steps and scale further with votes, whereas extra full renders improve slowly and expensively. On Sokoban, where errors correlate across seeds, sketches match full rendering at lower cost.}
    \label{fig:budget}
\end{figure*}

\Cref{tab:tiers} evaluates compute routing across 14 model-benchmark pairs.
No tier is best everywhere: direct answering wins 9 of the 14 pairs, including five of \tm's seven benchmarks (\eg, 71.7\% on VStar \vs 64.9\% Text and 58.1\% Draw), while cutting per-question latency on \tm to 0.3--1.1\,s (\vs 17--38\,s for Text and 38--73\,s for Draw).
Conversely, drawing leads on \vsp for \tm and \zebra, on ChartQA for \tm, and on VStar for \zebra, whose thought is causal there (52.9\% \vs 40.3\% without drawing, $p{=}0.009$), whereas text-only reasoning leads on \canvas.
Selecting a tier from 10 labeled questions reaches 65.1\% mean accuracy (64.6\% with 5 labels and 65.7\% with 20), gaining 4.4 points over default interleaved drawing and trailing the hindsight oracle by 2.3 points.
Although a fixed tier suits \zebra (Direct) or \canvas (Text) individually, that preferred tier flips across models, and 10-shot selection approaches each within 0.6 and 2.7 points without prior knowledge.
Unsupervised cross-tier agreement cannot replace these few labels: high agreement guarantees small accuracy gaps, whereas low agreement is ambiguous (\zebra's answers with and without drawing agree on only 46--79\% of items across benchmarks even when accuracy barely shifts).

\subsection{Plans become legible early}
\label{sec:exp_commit}

\Cref{fig:commit} tracks how spatial plans emerge over denoising steps.
At 1--2 steps, \tm emits a noisy copy of the input map without a legible path (\cref{fig:teaser}), matching \full on a third of \vsp puzzles, and at 4 steps commitment remains significantly below cross-seed agreement.
By 8 steps, the sketch commits to the exact plan of \full on 65.8\% of puzzles, exceeding cross-seed full-render agreement (61.2\%), while single-thought accuracy (69.7\%) is statistically indistinguishable from \full (72.2\%, $p{=}0.15$).
\wan behaves the same way: an 8-step video traces the same route as its 50-step counterpart on 69.4\% of mazes, matching cross-seed full-video agreement (72.2\%). It agrees equally often (69.4\%) with full videos from independent seeds, so the match does not come from the shared noise.
On \vrlvr, 8-step sketches likewise match same-seed full routes on 53.0\% of mazes, equal to cross-seed full-video agreement (53.0\%).
Beyond 8 steps, accuracy and commitment plateau as later steps refine visual texture.

As the step count drops from 50 to 2 on \wan, goal arrival falls from 80.0\% to 54.2\%, yet optimality \emph{among goal-reaching videos} stays at 90--93\%.
Under-denoised videos rarely draw a coherent wrong detour: the ball blurs, fragments, or stalls mid-corridor (\cref{fig:teaser}), and the sketch abstains at no cost to the vote.
\vrlvr's 8-step sketches similarly paint no valid route more often than full videos (38\% \vs 26\%) but commit to a wrong route less often when one appears (6\% \vs 13\%).
\tm's failed sketches mirror its full renders, with roughly half of erroneous paths stepping into a hole across step counts.

\begin{figure*}[t]
    \centering
    \includegraphics[width=\textwidth]{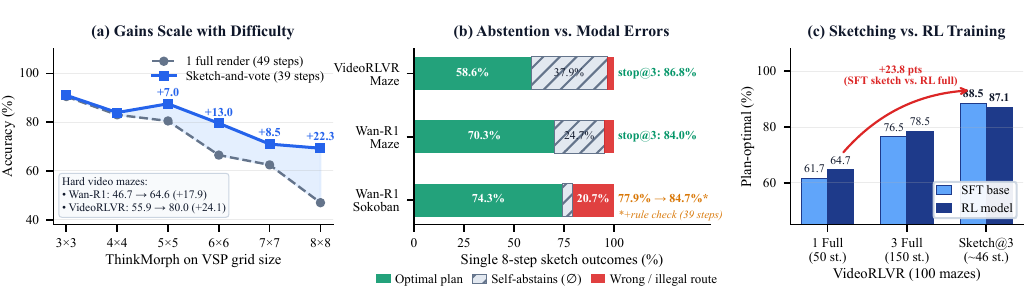}
    \caption{\textbf{Diagnostic analysis of sketch-and-vote.}
    \textbf{(a)} Gains grow monotonically with grid size and maze difficulty.
    \textbf{(b)} Single 8-step video outcomes: maze failures self-abstain ($\varnothing$), whereas Sokoban failures repeat illegal routes that a rule check filters out.
    \textbf{(c)} On \vrlvr (100 mazes), sketching on the SFT base model beats a 50-step rollout of the RL model by +23.8 points.}
    \label{fig:analysis}
\end{figure*}

\subsection{Sketch-and-vote}
\label{sec:exp_vote}

\Cref{tab:main,fig:budget} compare test-time allocations across models and tasks.
A single 8-step sketch trails a full render slightly but costs about one-sixth of the denoising steps, so a vote over a few sketches soon overtakes full rendering.
On \vsp, \stopm{3} draws roughly five sketches on average (39.1 steps) to reach 80.3\%, beating a 49-step full render by 8.6 points at 20\% lower step cost, whereas two full renders (98 steps) stall at 71.8\% due to random tie-breaking.
On \vrb mazes, \stopm{3} solves 84.0\% in 39 steps (+9.6 points over one full render) and matches or exceeds the five-full-render average on all 72 levels.
Because goal-reaching sketches rarely err, \stopm{2} already reaches 84.0\% in 29 steps, and \stopm{1} gains +6.6 points in 16 steps.
Abstention alone does not account for these gains: applying \stopm{1} to full renders gains only 4.9 points at 76 steps, half the gain of \stopm{3} at nearly twice its step budget.

\paragraph{Wall-clock time.}
Because each \wan video also pays fixed prompt-encoding and 193-frame VAE-decoding overhead, an 8-step sketch takes 48\,s \vs 167\,s for a 50-step video. \stopm{2} therefore matches single-render latency (176\,s) and keeps all +9.6 points, and \stopm{3} (234\,s) runs in under a third of the time of a 5-full-video vote (833\,s, which improves on zero mazes).
For \tm on 40 \vsp puzzles end-to-end, \stopm{3} adds roughly 10\% over a 39.4\,s full render, while \stopm{2} finishes faster than one full render.

\paragraph{Zero-shot transfer.}
Transferred unchanged to \vrlvr, \stopm{3} solves 86.8\% of mazes in 46 steps \vs 64.7\% for one full video (+22.1 points) and 79.0\% for a 3-full-video vote at 150 steps ($p{=}0.005$, with 32 mazes improved and 10 hurt).
Because incomplete paintings abstain, \stopm{1} already solves 84.8\% in 20 steps and 72\,s, beating a single 99\,s full video in both speed and accuracy (+20.1 points).
Under \vrlvr's noisy official check, \stopm{3} still gains +6.1 points (76.8\% \vs 70.7\%, $p{=}0.008$).
On a 100-maze subset, sketch-and-vote lifts \vrlvr's supervised SFT checkpoint to 88.5\% in 48 steps, outperforming a full rollout of the RL-trained model by 23.8 points (\cref{fig:analysis}c).
On Sokoban, sketch-and-vote matches full rendering in 32 \vs 50 steps (77.9\% \vs 79.2\%, $p{=}0.58$), and voting over three full videos also saturates (+1.3 points) because errors correlate across seeds (\cref{sec:analysis}).
On Style~2 mazes (15.7\% single-video accuracy), \stopm{3} matches three full renders (20.1\%) in under half their steps (71 \vs 150).

\subsection{Ablations}
\label{sec:exp_ablation}

\paragraph{Steps per draft.}
Near the budget of one full render, $k{=}8$ is optimal across all three reasoners: three 16-step drafts are too few to vote reliably, while single 4-step drafts drop 13--19 points below full renders.
Without labels, our commitment rule (\cref{sec:vote}) picks $k{=}8$ on \vsp, \vrb mazes (Style~1), Sokoban, and \vrlvr mazes, and $k{=}2$ on saturated trap fields. It under-selects ($k{=}2$) only in Style~2 mazes, where \wan fails on most levels so cross-seed agreement reflects shared errors rather than legibility.

\paragraph{Stopping threshold and scoring.}
Larger $m$ helps when individual sketches make non-abstaining errors: on \vsp, $m{=}2, 3, 4$ yield 78.7\%, 80.3\%, and 80.7\% at 26, 39, and 49 steps, whereas on both video maze tasks $m{=}2$ already matches $m{=}3$.
Cross-validating $m$ across stratified halves of \vsp within a 49-step budget picks $m{=}3$ or $m{=}4$, gaining +10.3 and +7.3 points out of sample ($p<0.001$).
Gains on \wan mazes hold under other metrics: without goal completion (\cref{sec:readout}), \stopm{3} beats full rendering 83.3\% to 66.4\%, and under \vrb's official SR 79.5\% to 70.3\% (both $p<0.01$).
On Sokoban, official SR (which fails 25\% of ground-truth videos while passing 42\% of full renders with illegal box routes) scores \stopm{3} at 64.7\% \vs 70.4\% ($p{=}0.05$).

\section{Analysis}
\label{sec:analysis}

\paragraph{Why votes over sketches succeed.}
Majority voting works when valid plans concentrate while errors disperse or self-abstain (\cref{fig:analysis}b).
On \vsp, wrong 8-step sketches scatter across paths: at least one of ten sketches solves 91.5\% of puzzles, and their plurality already recovers 80.8\%.
On \vrb mazes, label-free voting nearly matches oracle Pass@$K$ selection~\cite{yang2025vrbench} (84.0\% \vs 84.7\% over ten sketches, and 80.3\% \vs 81.9\% over five full videos) because incomplete rollouts abstain rather than splitting the vote.
Truncating to 8 steps preserves cross-seed diversity: two sketches share a plan on 55.7\% of \vsp puzzles and 65.9\% of \wan mazes, tracking full renders (61.2\% and 72.2\%).
Training a LoRA reader adapter on \tm's post-thought text adds only 1.2 points on \vsp (81.5\% \vs 80.3\% at \stopm{3}), so the off-the-shelf model already reads coarse sketches nearly as well as a trained reader.

\paragraph{When voting saturates: modal errors.}
On Sokoban, failed sketches rarely abstain (\cref{fig:analysis}b): 95\% push the box onto the target, a fifth via illegal paths (\vs $\sim$5\% wrong goal-reaching routes on mazes).
These errors repeat across seeds: given one wrong full render, another seed takes the same wrong route 55\% of the time (\vs 23\% on \wan mazes and 6\% on \vrlvr), so an illegal route wins $\ge 3$ of ten sketch votes on 14 of 72 levels (\vs 2 mazes).
Voting alone cannot overturn errors that repeat across seeds, but a solution-free rule check that turns illegal pushes into abstentions lifts this ceiling: \stopm{3} then solves 84.7\% of Sokoban levels in 39 steps (\vs 79.2\% for one full render, $p{=}0.04$), matching rule-filtered full renders at 69 steps.
Rejecting wall-crossing routes in the same way lifts \stopm{3} on Style~2 mazes to 23.6\% in 73 steps (\vs 15.7\% for one full video, $p{=}0.013$), an accuracy that rule-filtered full videos reach only at 132 steps.

\paragraph{Where gains concentrate.}
Extra samples add little where single renders already succeed, so gains concentrate on hard instances (\cref{fig:analysis}a): on \vsp, gains grow from +0.5 points on $3{\times}3$ grids to +22.3 on $8{\times}8$ grids (69.3\% \vs 47.0\%), and reach +17.9 and +24.1 points on hard \wan and \vrlvr mazes.

\paragraph{Sketches where drawing hurts.}
On VStar, where 49-step thoughts hurt \tm accuracy, truncating the same noise trajectory to 8 steps recovers nearly all of the loss (64.4\% \vs 64.9\% text-only and 58.1\% full, \cref{tab:regimes}b).
The sketch answers 15 questions that the full render misses and misses only 3 that it answers ($p{=}0.0075$).

\paragraph{Practical recipe.}
For a new model and task, first pick the compute tier on roughly ten labeled examples (\cref{sec:exp_tiers}), as direct answering is fastest and often most accurate.
When drawing wins, pick $k$ via our unlabeled commitment rule (\cref{sec:vote}, giving $k{=}8$ across \tm, \wan, and \vrlvr) and vote up to $m{=}3$, or $m{=}2$ when failed rollouts self-abstain, as unfinished maze videos do.
Voting helps as long as errors vary across seeds or can be caught by a solution-free rule check.

\section{Conclusion}
\label{sec:conclusion}

Generative visual reasoners inherit dense denoising schedules from photorealistic synthesis, as if visual polish were a prerequisite for reasoning.
Our interventions show that visual thoughts are often decorative or harmful, and a 10-shot choice among direct answering, text-only reasoning, and drawing adds 4.4 accuracy points on average while often cutting latency by an order of magnitude.
Where visual thoughts are causal, the symbolic plan is legible within 8 of the 49 or 50 denoising steps.
Spending the budget of one polished render on a few rough sketches that self-abstain and vote raises accuracy at lower denoising cost on both image and video reasoners.
Neither the router nor sketch-and-vote needs training. The router uses ten labeled examples per model and task, and sketch-and-vote picks its step count without labels and carries over unchanged from \wan to \vrlvr, where it adds 22.1 points over one full video.

{
    \small
    \bibliographystyle{ieeenat_fullname}
    \bibliography{main}
}

\end{document}